# On the Idea of a New Artificial Intelligence Based Optimization Algorithm Inspired From the Nature of Vortex

*Utku Kose*
Computer Sciences Application and Research Center
Usak University, Usak, Turkey
utku.kose@usak.edu.tr

*Ahmet Arslan*
Department of Computer Engineering
Selcuk University, Konya, Turkey
ahmetarslan@selcuk.edu.tr

**Abstract**

In this paper, the idea of a new artificial intelligence based optimization algorithm, which is inspired from the nature of vortex, has been provided briefly. As also a bio-inspired computation algorithm, the idea is generally focused on a typical vortex flow / behavior in nature and inspires from some dynamics that are occurred in the sense of vortex nature. Briefly, the algorithm is also a swarm-oriented evolutional problem solution approach; because it includes many methods related to elimination of weak swarm members and trying to improve the solution process by supporting the solution space via new swarm members. In order have better idea about success of the algorithm; it has been tested via some benchmark functions. At this point, the obtained results show that the algorithm can be an alternative to the literature in terms of single-objective optimization solution ways. Vortex Optimization Algorithm (VOA) is the name suggestion by the authors; for this new idea of intelligent optimization approach.

## 1. Introduction

Rapid developments and improvements within different technologies have remarkable roles on improving our modern life and providing effective solutions to the real-world problems. At this point, it is always omitted that mathematical dynamics of the related solution approaches are generally based on natural dynamics. In other words, it is clear that nature – biological environment has many features and functions that can be inspired from in order to have novel solution approaches for especially, hard, complex, real-world based problems [1, 2]. When we consider mathematical solution approaches, methods, and techniques, it can be seen that the nature has a big role on forming each formulations in the sense of providing alternative solutions. Even simple mathematical equations have some bio-inspired sides that researchers / scientists have inspired from during designing these mathematical systems. In this sense, we can briefly say that nature and biological world has a big role on thinking about solutions and designing mathematical structures in the way of developing effective computational approaches, methods, or techniques. At this point, the optimization concept has been widely inspired from the nature for many years; as one of the related mathematical solution approaches.

When we take the optimization concept into consideration, we can see that it has a great importance on research interests in the intersection of different literatures associated with problems covering optimization solutions and alternative approaches in order to satisfy needs in typical optimization solution ways. In time, the related optimization approaches have gained a remarkable momentum in designing novel methods, and techniques for being alternative to real-world based optimization problems. As general, the associated literature has especially many different algorithms inspired from behaviors of different organisms or natural dynamics while designing mathematical infrastructure, which is strong enough in order to cover optimization problems and provide effective solutions for them. At this point, swarm intelligence is a remarkable research interest, which has a great role on providing effective solutions for optimization operations and has strong relation with bio-inspired computation [3–7]. It seems that the future of bio-inspired computation will be greatly affected by such developments and gain rapid improvement flow as the number of different problems increases in time. On the other hand, we can say that the future of artificial intelligence will be shaped from not only inspirations on human thinking / behavior approaches but also from inspirations on natural dynamics.

Objective of this paper is to introduce the idea of a new artificial intelligence based optimization algorithm, which is inspired from the nature of vortex. As also a bio-inspired computation algorithm, the idea is generally focused on a typical vortex flow / behavior in nature and inspires from some dynamics that are occurred in the sense of vortex nature. From a general perspective, the algorithm is also a swarm-oriented evolutional problem solution approach; because it includes many methods related to elimination of weak swarm members and trying to improve the solution process by supporting the solution space via new swarm members. It also employs simple mathematical equations in order to be formed and applied easily in optimization problems. In order have better idea about success of the algorithm; it has been tested via some benchmark functions and the results received from the tests show that the algorithm can be an alternative to the literature in terms of single-objective optimization solution ways. Vortex Optimization Algorithm (VOA) is the name suggestion by the authors; for this new idea of intelligent optimization approach.

In the context of the objective of this paper, the remaining content is organized as follows: The next section is devoted a brief look at to the history of the idea and the development process so far. After that, the fourth section provides the fundamentals of the designed algorithm. It briefly explains the approach and provides the general structure of the algorithm, which is also called as the Vortex Optimization Algorithm (VOA). Next to the fourth section, some brief test / evaluation processes performed via optimization benchmark functions are reported under the fifth section. Finally, the paper is ended by providing conclusions and discussing about future works.

## 2. The History of the Idea / Development Process

Before explaining the idea in detail and shaping it in an algorithm form, it is a good idea to focus on the background related to the idea and the development process so far. Briefly, we can call this background information as a brief history of the idea / development process. In this sense, the foremost points regarding to 'the history' can be listed as follows:

First ideas regarding to usage of vortex behaviors for optimization approaches have appeared when the authors had the following experiences / examinations in terms of interactions with the nature / real-world:

1. Vortex flow appeared in water when the plug hole is opened.
2. Vortex flows created by the passage of plane wing or by an engine of a plane.
3. Vortex shapes appeared in the nature / space; because of different environmental conditions.

After having ideas to form a solution approach for optimization problems, there has been a need for employing some 'intelligent mechanisms' in order to have effective solution steps based on the power of the artificial intelligence.

The idea of forming an artificial intelligence based optimization algorithm has been shaped in terms of swarm-intelligence and also evolutionary computing. In other words, the solution approach has been a swarm-oriented, evolutional one.

Some dynamics from vortex nature has been tried to be applied for the swarm members of the algorithmic approach. At this point, the following applications has been done:

1. Real mathematical approaches regarding to different vortex types (like rigid-body or irrational vortex) has been tried to be used for movement of swarm members within the solution space. But this application has caused the algorithm structure to be complex and not effective enough for optimization operations.

2. As an alternative, typical spiral move has been applied for the movement of swarm members; by using the appropriate mathematical equations. But as similar to the previous experience, this application has also been not effective enough for the related optimization operations.

3. Because of the previous experiences, the authors have decided to use some simple mathematical equations, which are not directly based on certain equations of vortex dynamics but inspired from the nature of vortex. In this sense, it has been aimed to form a simple mathematical structure that can be formed and applied easily.

The last application regarding to the related mathematical equations has been improved after some testing processes and the first stable form of the algorithm has been obtained. It is important to indicate according to the last form of the algorithm that each swarm member (particle) has 'vorticity' value, which is changed / updated along the solution steps. At this point, vorticity values affect the fitness values obtained with objective functions. Of course, positions of the particles are also changed / updated under the sway of particle vorticity values, positions / vorticity values regarding to the best particles...etc. As addition to the employment of swarm-intelligence, this algorithmic structure has provided an evolutionary approach, which aims promoting the particles by making their status 'vortex' or 'normal' particle and removing normal (weak) members from the solution space in order to immediately support the general swarm group with new members, which are 'ready to rock'. This mechanism is achieved by using an elimination value, which is a border for applying the related operation.

After having at least a first, stable form of the algorithm, which is based on the nature of vortex, it has been possible to draw the general structure of the solution steps. In this sense, the next section is for a brief report of the algorithmic structure and its working mechanism.

### 3. New A.I. Based Optimization Algorithm Inspired from the Nature of Vortex

As it was also indicated before, the algorithm is generally focused typical vortex flow – behavior in nature and inspires from some dynamics that are occurred in the sense of vortex nature. At this point, it is important to report that the authors have decided to call this algorithm as the Vortex Optimization Algorithm (VOA). Algorithmic details of the VOA can be expressed briefly as follows:

1. **Step 1:** Define initial parameters ($N$ for number of particles; initial *vorticity* ($v$) values of each particle; max. and min. limits (min. limit is the negative of the max. one) for vorticity value ($max\_v$ and $min\_v$) and other values related to function, problem…etc. (like dimension, search domain…etc.); and finally $e$ for the elimination rate.

2. **Step 2:** Locate the particles randomly within the solution space and calculate fitness values for each of them. Update the $v$ value of the particle with the best fitness value by using a random value (Equation 1). Mark this particle as a 'vortex' and keep its values as the best one so far.

3. *the_best_particle_at_first_v_(new) = the_best_particle_at_first_v_(current) +*

4. *(random_value * the_best_particle_at_first_v_(current))* (1)

5. **Step 3:** Repeat the sub-steps below in the sense of the stopping criteria (for example iteration number):

   a. **Step 3.1:** Mark each particle, whose fitness value is equal to or under the average fitness of all particles (if the problem is minimization), as the 'vortex'. The other / remaining particles are in the 'normal' particle status.

   b. **Step 3.2:** Update $v$ value of each particle ($i$) by using the following equations:
   *particle_i_v_(new) = particle_i_v_(current) + (random_value * (global_best_v / particle_i_v_(current)))* (2)

   c. **Step 3.3:** Update the $v$ value of each vortex particle (except from the best particle so far) by using a random value (Equation 3).
   *particle_i_v_(new) = random_value * particle_i_v_(current)* (3)

   d. **Step 3.4:** Update position of each particle (except from the best particle so far) by using the following equation:

$particle_i\_position\_(new) = particle_i\_position\_(current) + (random\_value *$
$(particle_i\_v\_(current) * (global\_best\_position -$
$particle_i\_position\_(current))))$         (4)

e. **Step 3.5:** Calculate fitness values according to new positions of each particle. Mark the particle with the best value as a 'vortex' (if it is not a vortex yet) and keep its values as the best so far.

f. **Step 3.6:** If number of non-vortex particles is equal to or under the value of *e*, remove all non-particles from the solution space and create new particles according to number of removed particles. Locate these new particles randomly within the solution space. Return to the Step 3.1. if the stopping criteria has not been reached yet.

6. **Step 4:** The best values obtained within the loop is the optimum solution.

As it can be seen from the algorithm steps, the VOA employs simple equations. It is an advantage that the algorithm can be formed and applied within optimization problems whereas alternative algorithms may contain some complex solution steps (This situation may be also an disadvantage for the VOA when it is applied in more difficult optimization problems but while the world is transformed into a 'strong simplicity', the VOA may be a practical solution approach).

The working mechanism of the VOA can be visualized briefly as like in Figure 1.

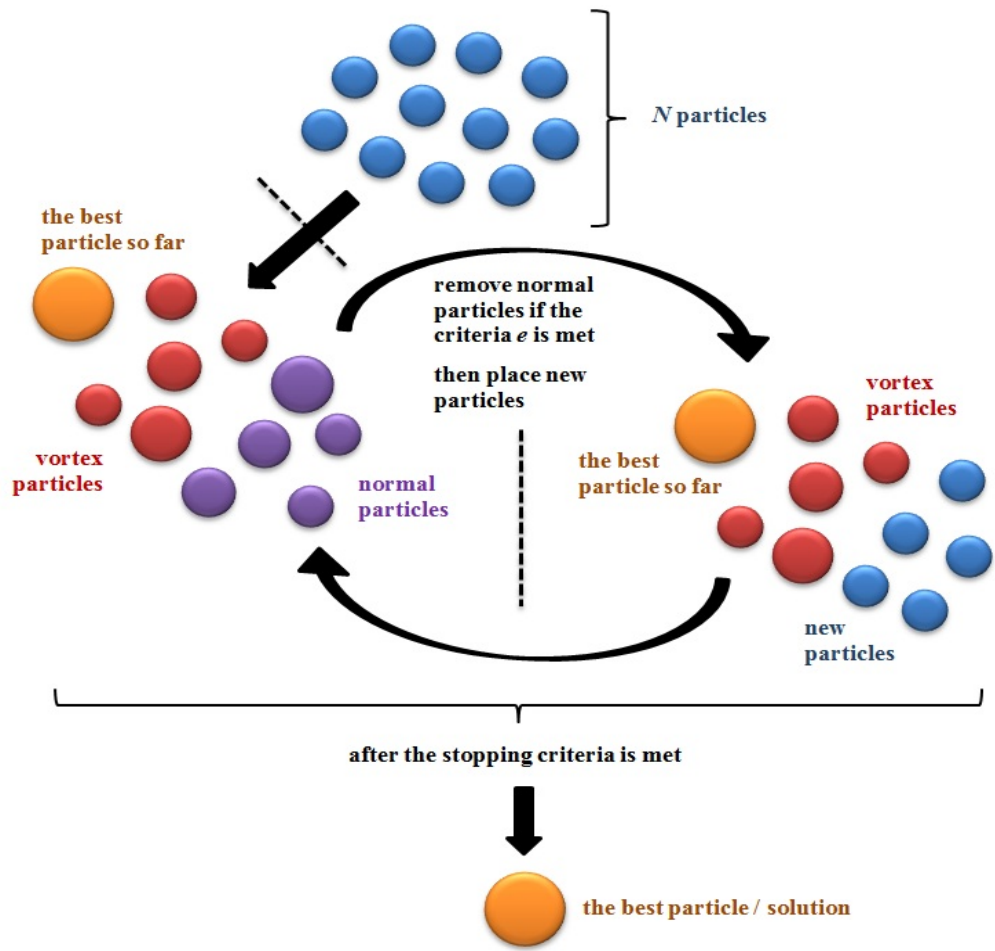

Figure 1. Working mechanism of the VOA.

63

## 4. Evaluation

In order to understand effectiveness of the VOA and evaluate its success, some single-objective optimization benchmark functions have been solved in terms of different dimensions. General conditions for the performed tests are as follows:

- Total number of particles ($N$): 50
- Total iteration (the stopping criteria): 5000
- Initial vorticity value: 0.50
- Max. vorticity value: 7.0
- Min. vorticity value: -7.0)
- Elimination rate ($e$): 50

- Dimensions: 2 for first four functions and 2, 5, 10, 20, and 30 respectively for the last two functions.
- Other specific values have been determined according to characteristics of each function.

Table 1 provides a list of information regarding to the benchmark functions that have been used along to the test process.

Table 1. Benchmark functions that have been used along to the test process.

| Function | Formula | Search Domain | Minimum |
|---|---|---|---|
| Booth's | $f(x,y) = (x + 2y - 7)^2 + (2x + y - 5)^2$ | $-10 \leq x, y \leq 1$ | $f(1, 3) = 0$ |
| Beale's | $f(x,y) = (1.5 - x + xy)^2 + (2.25 - x + xy^2)^2 + $ | $-4.5 \leq x, y \leq$ | $f(3, 0.5) = 0$ |
| Goldstein–Price | $f(x,y) = (1 + (x + y + 1)^2 (19 - 14x + 3x^2 - 14y$ | $-2 \leq x, y \leq 2$ | $f(0, -1) = 3$ |
| McCormick | $f(x,y) = sin(x + y) + (x - y)^2 - 1.5x + 2.5y + 1$ | $-1.5 \leq x \leq 4$ $-3 \leq y \leq 4$ | $f(-0.54719, -1.5$ |
| Three-hump camel | $f(x,y) = 2x^2 - 1.05x^4 + \frac{x^6}{6} + xy + y^2$ | $-5 \leq x, y \leq 5$ | $f(0,0) = 0$ |
| Sphere | $f(x) = \sum_{i=1}^{n} x_i^2$ | $-100 \leq x_i \leq 10$ $1 \leq i \leq n$ | $f(x_1, ..., x_n) = f(0$ |
| Rosenbrock | $f(x) = \sum_{i=1}^{n-1} [100(x_{i+1} - x_i^2)^2 + (x_i - 1)^2]$ | $-30 \leq x_i \leq 30$ $1 \leq i \leq n$ | $f(x_1, ..., x_n) = f(1$ |

Table 2 provides a brief report related to the obtained results after the related tests / operations over the benchmark functions.

As it can be seen from the Table 2, the VOA reaches to the desired values for the related functions. Although the number of total iteration and also particles were low (5000 and 50) the VOA can be closer to minimum values especially for bigger dimensional Sphere and Rosenbrock functions.

Table 2. A brief report related to the obtained results.

| Function | Minimization Results for the Dimensions as: | | | | |
|---|---|---|---|---|---|
| | 2 | 5 | 10 | 20 | 30 |
| Booth's | 0.0000 | x | x | x | x |
| Beale's | 0.0000 | x | x | x | x |
| Goldstein–Price | 3.0000 | x | x | x | x |
| McCormick | -1.9133 | x | x | x | x |
| Three-hump camel | 0.0000 | x | x | x | x |
| Sphere | 0.0000 | 0.0000 | 0.0000 | 0.0000 | 0.0000 |
| Rosenbrock | 0.0000 | 0.0000 | 0.0002 | 0.0027 | 0.0023 |
| x: not applicable | | | | | |

### 5. Conclusions and Future Work

This paper has been an introduction for the idea of a new artificial intelligence based optimization algorithm, which is inspired from the nature of vortex. In this sense, the paper has provided information about how the idea has been realized with an algorithmic structure and reported brief test / evaluation processes that were done via the formed algorithm. Called as the Vortex Optimization Algorithm (VOA), the approach seems to effective enough for solving single-objective optimization problems. One of the most important advantages of the VOA is its simple mathematical infrastructure and as an alternative solution approach, the VOA can be formed and applied easily.

Future works regarding to the VOA can be explained briefly as follows:

- In the sense of test / evaluation processes, the paper has focused on some single-objective benchmark functions. So, there will be more work on other remaining benchmark functions.
- Further works will be based on also comparison of the VOA with other alternative algorithms in the literature.
- The authors have already started some works in order to solve different optimization related problems from different fields.
- Additionally, there will be also more works on evaluation of VOA particle / algorithm parameters; in order see if changes in the related values can improve performance and accurateness.